\documentclass{article} % For LaTeX2e
\pdfoutput=1  %% Necessary only when you compile this at arXiv
\usepackage{iclr2015,times}
\usepackage{url}

\usepackage{graphicx}
\usepackage{amsmath}
\usepackage{amssymb}
\usepackage{ascmac}

\usepackage{algorithm,algpseudocode} % pseudo code: algorithmicx package

\title{A Bayesian encourages dropout}

\author{
Shin-ichi~Maeda % \& Shin Ishii % \thanks{ Use footnote for providing further information
\\
Graduate School of Informatics\\
Kyoto University\\
Yoshidahonmachi 36-1, Sakyo, Kyoto, Japan \\
\texttt{ichi@sys.i.kyoto-u.ac.jp}
}

\iclrconference % Uncomment if submitted as conference paper instead of workshop

\begin{document}

\maketitle

\begin{abstract}
Dropout is one of the key techniques to prevent the learning from overfitting.
It is explained that dropout works as a kind of modified L2 regularization.
Here, we shed light on the dropout from Bayesian standpoint.
Bayesian interpretation enables us to optimize the dropout rate, 
which  is beneficial for learning of weight parameters
and prediction after learning. 
The experiment result also encourages the optimization of the dropout.
%
%According to the Bayesian standpoint, we can consider that the dropout
%performs an approximate Bayesian learning and inference,
%and this view enables us to optimize the dropout rate.
%The optimization of the dropout rate is beneficial for weight parameters
%and prediction. The experiment result also encourages the optimization of the dropout.
\end{abstract}

\section{Introduction}
Recently, large scale neural networks has successfully shown its great 
learning abilities for various kind of tasks such as
image recoginition, speech recognition or natural language processing.
Although it is likely that the large scale neural networks
fail to learn the appropriate parameter because of its huge model complexity,
the development of several learning techniques, optimization 
and utilization of large dataset overcome such difficulty of learning.

One such successful learning technique is `dropout', 
which is considered to be important to suppress the overfitting~\citep{Hinton12dropout}.
The success of dropout attracts many researchers' attention
and several theoretical studies have been performed aiming 
for better understanding of dropout~\citep{Wager13regularizeddropout, Baldi13regularizeddropout, Ba13regularizeddropout}.
The most of them explain the dropout as a kind of new input invariant regularization technique,
and prove the applicability to other applications such as generalized linear model~\citep{Wager13regularizeddropout, Wang13fastdropout}.

In contrast, we formally provide a Bayesian interpretation to dropout.
From this standpoint, we can think that the dropout
 is a way to solve the model selection problem
 such as input feature selection or the selection of number of hidden internal states
 by Bayesian averaging of the model.
That is, the inference after dropout training can be considered as an 
approximate inference by Bayesian model averaging where each model is weighted in accordance with the posterior distribution,
and the dropout training can be considered as an approximate learning of parameter which optimizes a weighted sum 
of likelihoods of all possible models, i.e., marginal likelihood.
This interpretation enables the optimization of `dropout rate' as 
the learning of the optimal weights of the model, i.e., posterior of the model.
This optimization of dropout rate benefits
the parameter learning by a better approximation of marginal likelihood 
and prediction by a closer distribution to the predictive distribution.
Note the Bayesian view on `dropout' was already seen in the original paper of dropout~\citep{Hinton12dropout}
which regards the dropout training as a kind of approximate Bayesian learning,
however it was not mentioned how much the gap would be %
between the dropout training and standard Bayesian learning,
and how to set or optimize the dropout rate.

%The paper is organized as follows;
%Chapter 2 explains standard dropout algorithm,
%and Chapter 3 explains the existing interpretations of the dropout.
%Then the Bayesian interpretation of dropout is presented in Chapter 4.
%Chapter 5 shows the experiment which validates the effectiveness 
%of the optimization of dropout rate.
%Finally Chapter 6 is devoted for the disscussion.

%近年、大規模なニューラルネットワークの学習が成功をおさめるようになった。
%大規模なモデルは過学習のおそれがあるが、Dropoutはその過学習を防ぐのに重要な技術とみなされている。
%なぜうまくいくのかの理論付けとともにニューラルネットワーク以外への応用も多くなされてきた[文献]
%
%本論文では、従来とはdropoutに対してBayesian inferenceからのviewを提供する。
%この視点にたつことで、dropout rateの最適化を可能とする。
%このdropout rateの最適化は、より望ましいdesirable cost function、すなわち
%marginal likelihoodにより近いコスト関数でのパラメータの学習をencourage？促し、 
%より予測分布に近い分布に基づいた予測を可能とする。
%
%これらの解析にもとづいた簡単な実験により、Bayesian dropoutのStandard dropoutに対する
%優位性を述べる。
%
%Optimal model mixture
%Bayesian explains dropout

\section{standard dropout algorithm} \label{sec:notation}
\subsection{Dropout training} \label{sec:dropout training}
We will explain the dropout for three-layered neural network 
not only because the neural network is the first model that the dropout training is applied,
but also because it is one of the simplest models that involves both of the
 input feature selection problem and the hidden internal units selection problem.
Note the concept of dropout itself can be easily 
extended to any other learning machines.

Let $\mathbf{x} \in {\Re ^n}$, $\mathbf{h} \in {\Re ^m}$,
and $\mathbf{y} \in {\Re ^l}$ be $n$-dimensional input, 
$m$-dimensional hidden unit, and $l$-dimensional output, respectively.
Then, the hidden unit $\mathbf{h}$ and the output $\mathbf{y}$ are computed as follows;
\begin{align}
 \begin{cases}
{\mathbf{h}} &= \sigma (W^{(1)}\mathbf{x} + \mathbf{b}^{(1)}), \\ 
 \mathbf{y} &= \sigma (W^{(2)}\mathbf{h} + \mathbf{b}^{(2)}),
\nonumber %\label{chap3:eq:infinite_Gibbs}
 \end{cases}
\end{align}
where $W^{(1)}$ and $W^{(2)}$ are $m \times n$ and $l \times m$ matrix, 
and  $\mathbf{b}^{(1)}$ and $\mathbf{b}^{(2)}$ are $m$- and $l$-dimensional vector, respectively.
$\sigma(\cdot )$ is a nonlinear activation function such as a sigmoid function.
The set $\theta = \{W^{(1)}, W^{(2)}, \mathbf{b}^{(1)}, \mathbf{b}^{(2)} \}$ is a parameter set to be optimized.

Now we consider the model selection problem
where the full model is represented as the one described above,
and the other possible models (submodels) consist of the subset of the
input features or the hidden units of the full model.
Let us introduce the diagonal mask matrices  $Z^{(1)}$ and $Z^{(2)}$ 
where the $i$-th diagonal elements $z^{(1)}_i \in \{0, 1\}$ $(i=1,\cdots,n)$ and
$z^{(2)}_j \in \{0, 1\}$ $(j=1,\cdots,m)$ are both binary to represent such subsets.
We will sometimes use $\mathbf{z} = \{z^{(1)}_i, z^{(2)}_j | i=1,\cdots,n, j=1,\cdots,m \}$ 
% $\mathbf{z} = \{\rm{diag}(Z^{(1)}), \rm{diag}(Z^{(2)}) \}$ 
to represent all the mask variables. %
Then the possible model is represented as follows;
\begin{align}
 \begin{cases}
{\mathbf{h}} &= \sigma (W^{(1)}Z^{(1)}\mathbf{x} + \mathbf{b}^{(1)}), \\ 
 {\mathbf{y}} &= \sigma (W^{(2)}Z^{(2)}\mathbf{h} + \mathbf{b}^{(2)}).
\nonumber %\label{chap3:eq:infinite_Gibbs}
 \end{cases}
\end{align}
The determination of $Z^{(1)}$ corresponds to the input feature selection problem
while the determination of $Z^{(2)}$ corresponds to the hidden internal units selection problem.
We may rewrite ${\mathbf{y}}({\mathbf{x}};\theta )$ with
 ${\mathbf{y}}({\mathbf{x}};Z^{(1)}, Z^{(2)}, \theta )$
to represent the explicit dependence on the masks $Z^{(1)}$ and $Z^{(2)}$ 
as well as the parameter $\theta $.
There are $2^{(m+n)}$ possible architectures when
ignoring the redundancy brought by the hierarchy of the architecture,
and this number becomes huge in practice. 
This poses a great challenge in the optimization of the best mask $\mathbf{z}$, i.e., the best model selection.

Instead of choosing a specific binary mask $\mathbf{z}$,
the standard dropout takes the approach to stochastically mix all the possible models
by a stochastic optimization.
The algorithm is summarized as follows;\\
% \begin{itembox}[l]{ \textbf{ Standard dropout for learning }}
\fbox{\parbox{\textwidth}{\textbf{Standard dropout for learning }:
\begin{enumerate}
\item Set $t=0$ and set an initial estimate for parameter $\theta _0$.  \par 
\item Pick a pair of sample $({\mathbf{x}}_t, {\mathbf{y}}_t)$ at random. \par 
\item Randomly set the mask $\mathbf{z}_t$ %$Z^{(1)}_t$ and $Z^{(2)}_t$
         %by determining every diagonal element of the matrices %$z^{(1)}_i$ and $z^{(2)}_j$ %%
         by determining every element independently according to the dropout rate, typically, $\rm{Ber}(0.5)$
         where $\rm{Ber}(p)$ denotes a Bernoulli distribution with probability $p$. \par
\item Update the parameter $\theta $ as  \par
\begin{eqnarray}
 \theta _{t+1} = \theta _t + \eta _t \frac{\partial \log p(\mathbf{y}_t|\mathbf{x}_t, \mathbf{z}_t, \theta)}{\partial \theta },
 \nonumber 
\end{eqnarray} \par%
where $\eta _t$ determines a step size, and should be decreased properly to assure the convergence. %~\citep{Robbins51StocahsticGradient}.
In case of the training of the three-layered neural network for regression, 
$\log p(\mathbf{y}_t|\mathbf{x}_t, \mathbf{z}_t, \theta) \propto - ||{\mathbf{y}}_t - {\mathbf{y}}({\mathbf{x}}_t;Z^{(1)}_t, Z^{(2)}_t, \theta )||^2 _2$
where $||\cdot ||_2$ denotes a Euclidean norm. \par
%In case of the training of the three-layered neural network for regression, the above parameter update would be \par
%\begin{eqnarray}
% \theta _{t+1} = \theta _t - \eta _t \frac{\partial ||{\mathbf{y}}_t - {\mathbf{y}}({\mathbf{x}}_t;Z^{(1)}_t, Z^{(2)}_t, \theta )||^2 _2}{\partial \theta }
% \nonumber 
%\end{eqnarray} \par%
%where $||\cdot ||_2$ denotes Euclid norm.\par
\item Increment $t$ and go to step 2 unless certain termination
condition is not satisfied.
\end{enumerate}
}}
% \end{itembox}

\subsection{Prediction after dropout training}  \label{sec:dropout prediction}
After training of $\theta $, the output for the test input ${\mathbf{x}}^*$ is given as %
\begin{align}%
E_{p(Z^{(1)})p({Z^{(2)}})} \left[{\mathbf{y}}({\mathbf{x}}^{*};Z^{(1)}, Z^{(2)}, \theta ) \right] 
\approx {\mathbf{y}}\left({\mathbf{x}}^{*};E_{p(Z^{(1)})}[Z^{(1)}], E_{p(Z^{(2)})}[Z^{(2)}], \theta \right),
\end{align}
where $E_{p(\mathbf{z})}[f(\mathbf{z})]$ denotes an expected value of function $f(\mathbf{z})$ with respect to the distribution $p(\mathbf{z})$.
The distribution $p(Z^{(1)})$ and $p({Z^{(2)}})$ in this case must correspond to the ones used 
in choosing mask matrices $Z^{(1)}$ and $Z^{(2)}$ during learning.
If an independent Bernoulli distirbution with probability $p=0.5$ is used for both choosing $Z^{(1)}$ and $Z^{(2)}$,
then the three-layered neural network output ${\mathbf{y}}\left({\mathbf{x}}^{*};E_{p(Z^{(1)})}[Z^{(1)}], E_{p(Z^{(2)})}[Z^{(2)}], W^{(1)}, W^{(2)}, \mathbf{b}^{(1)}, \mathbf{b}^{(2)} \right)$
is equivalent to the output using halved weights ${\mathbf{y}}\left({\mathbf{x}}^{*};\frac{1}{2}W^{(1)}, \frac{1}{2}W^{(2)}, \mathbf{b}^{(1)}, \mathbf{b}^{(2)} \right)$.

There is another way to approximate $E_{p(Z^{(1)})p({Z^{(2)}})} \left[{\mathbf{y}}({\mathbf{x}}^{*};Z^{(1)}, Z^{(2)}, \theta ) \right]$
known as `fast dropout' ~\citep{Wang13fastdropout}.
When we apply dropout,
not only to neural network, but also to other models, 
we often need to compute a weighted sum of many independent random variables $\mathbf{w}^T Z {\mathbf{x}}$
where the upper script $T$ denotes a transpose of vector (or matrix),
$\mathbf{w}$ is a weight vector, and $Z$ is a  diagonal mask matrix whose diagonal elements are binary random variables.
${\mathbf{x}}$ is either a sample of the input or the hidden unit, and treated as a fixed value.
Fast dropout approximates this weighted sum of random variables by a single Gaussian random variable 
utilizing Lyapunov's central limit theorem.
This Gaussianization improves the accuracy of the calculation of the expectation while saving the computation time.
% approximation %it is shown that this approximation significantly improves the test error while saving the computation time.
It is also shown that the Gaussian approximation is beneficial to the learning.

\section{Existing interpretations of dropout algorithm} \label{sec:existing methods}
Although the idea of dropout originates from Bayesian model averaging,
it is proposed that the artificial corruption of the model by dropout is
interpreted as a kind of adaptive regularization~\citep{Wager13regularizeddropout, Wang13fastdropout} 
analogous with the artificial feature corruption that is interpreted 
as a kind of regularization.
Under this viewpoint, the dropout regularizer used for generalized linear model 
can be seen as the first-order equivalent to $L_2$-regularization after transforming
the input by an estimate of the inverse diagonal Fisher information matrix~\citep{Wager13regularizeddropout}.
This input transformation let the regularizer scale invariant while the conventional
$L_2$-regularizer does not hold this desirable property.

In~\citep{Baldi13regularizeddropout}, %
the cost function used in dropout training is analyzed.
The cost function used in dropout training is considered as an average of the cost function of submodels.
They analyze the difference of the average of the cost function of submodels
and a cost function of the average of the submodels,
and show that the dropout brings an input-dependent regularization term to 
the cost function of the average of the submodels.
They also pointed out that the strongest regularization effect is obtained
when the dropout rate is set to be 0.5, which is a typical value used in practice.

There is also a study that treats dropout from Bayesian viewpoint.
\citet{Ba13regularizeddropout} extend the dropout to `standout'
 where the dropout rate is adaptively trained. 
They propose that the random masks used in dropout training should be considered 
as a random variable of the Bayesian posterior distribution over submodels,
which shares our view to the dropout.
However, their proposed adaptive dropout rate depends on each input variable,
which implies there should be a different mask for each input,
while here we assume a consistent mask for whole dataset, and
tries to estimate the best mix of models based on the entire training dataset.
Also, it is not clear why their update of the dropout rate works as the leaning of the posterior.
In the following section, we will show a clear interpretation of dropout from Bayesian standpoint,
and provide a theoretically solid way to optimize the dropout rate.
The difference of our study from the existing studies 
will be further discussed in Section~\ref{sec:discussion}.

\section{Bayesian interpretation of dropout algorithm} \label{sec:proposed method}
\subsection{Bayesian interpretation of dropout training}
In standard Bayesian framework, all the parameters are treated 
as random variables to be inferred via posterior except for the hyper parameter.
In dropout training, $\theta $ in Section~\ref{sec:notation} corresponds to the hyper parameter 
%although their dimensions are usually large in practice
while the mask $\mathbf{z}$ in Section~\ref{sec:notation} corresponds to the parameter.
%matrices $Z^{1}$ and $Z^{2}$ in Section~\ref{sec:notation} correspond parameters.
%To describe the dropout training in more general form,
Let us denote $\mathbf{z}$ an $m$-dimensional binary mask vector whose elements take either $0$ or $1$,
$D =\{\mathbf{x}_1, \mathbf{y}_1, \cdots, \mathbf{x}_T, \mathbf{y}_T \}$ the set of training data set.

Then the marginal log-likelihood for hyperparameter $\theta $ is defined as
\begin{align}
 {\log p(D|\theta )}
\equiv  \log \sum\nolimits_{\mathbf{z}} {\left( {\prod\nolimits_{t = 1}^T 
{p({\mathbf{y}_t}|{{\mathbf{x}}_t},{\mathbf{z}},\theta )} } \right)}p({\mathbf{z}}).
\nonumber 
\end{align}
Although we omit the dependence of $\theta $ on $p({\mathbf{z}})$ here for simplicity,
it is possible to include such dependence in general.
By introducing any distribution of the parameter ${\mathbf{z}}$, $q({\mathbf{z}})$,
which we call trial distribution following the conventions,
the marginal log-likelihood can be written as 
\begin{align}
& {\log p(D|\theta )} \nonumber \\%
=&  E_{q(\mathbf{z})}[\log p(D|\theta )] \nonumber \\
%\sum\nolimits_{\mathbf{z}} {q({\mathbf{z}})\log p(D|\theta )} \nonumber \\
=&  E_{q(\mathbf{z})}\left[\log \frac{{p(D,{\mathbf{z}}|\theta )}}{{p({\mathbf{z}}|D,\theta )}}\right] \nonumber \\
=&  E_{q(\mathbf{z})}[\log {p(D,{\mathbf{z}}|\theta )}] - E_{q(\mathbf{z})}[\log {q({\mathbf{z}})}] + KL\left[ {q({\mathbf{z}})|p({\mathbf{z}}|D,\theta )} \right] \nonumber \\
%=& \sum\nolimits_{\mathbf{z}} {q({\mathbf{z}})\log \left( {\left( 
%{\prod\nolimits_{t = 1}^T {p({\mathbf{y}_t}|{{\mathbf{x}}_t},{\mathbf{z}},\theta )} } \right)
%p({\mathbf{z}} )} \right)}  - \sum\nolimits_{\mathbf{z}} {q({\mathbf{z}})\log q({\mathbf{z}})} 
% + KL\left[ {q({\mathbf{z}})|p({\mathbf{z}}|D,\theta )} \right] \nonumber \\
\geq&  E_{q(\mathbf{z})}[\log {p(D,{\mathbf{z}}|\theta )}] - E_{q(\mathbf{z})}[\log {q({\mathbf{z}})}]  \nonumber \\
%\geq & \sum\nolimits_{\mathbf{z}} {q({\mathbf{z}})\log \left( {\left( 
%{\prod\nolimits_{t = 1}^T {p({\mathbf{y}_t}|{{\mathbf{x}}_t},{\mathbf{z}},\theta )} } \right)
%p({\mathbf{z}} )} \right)}  - \sum\nolimits_{\mathbf{z}} {q({\mathbf{z}})\log q({\mathbf{z}})}  \nonumber \\
=& \sum\nolimits_{t=1}^T \sum\nolimits_{\mathbf{z}} q({\mathbf{z}})\log p({\mathbf{y}_t}|{{\mathbf{x}}_t},{\mathbf{z}},\theta )
+ \sum\nolimits_{\mathbf{z}} q({\mathbf{z}}) \log p({\mathbf{z}})  - \sum\nolimits_{\mathbf{z}} {q({\mathbf{z}})\log q({\mathbf{z}})} 
\equiv F(q({\mathbf{z}}), \theta), \label{eq:lowerbound}
\end{align}
where $KL\left[q(\mathbf{z})|p(\mathbf{z}) \right] \equiv 
\sum\nolimits_{\mathbf{z}} q(\mathbf{z}) \log \frac{q(\mathbf{z})}{p(\mathbf{z})}$ 
denotes a Kullback-Leibler divergence between the distritbuions $q(\mathbf{z})$ and $p(\mathbf{z})$,
and $p({\mathbf{z}}|D,\theta ) \propto  \sum\nolimits_{\mathbf{z}} {\left( {\prod\nolimits_{t = 1}^T 
{p({\mathbf{y}_t}|{{\mathbf{x}}_t},{\mathbf{z}},\theta )} } \right)}p({\mathbf{z}}) $ is a posterior distribution.
The lower bound in Eq. (\ref{eq:lowerbound}) is valid for any trial distribution $q(\mathbf{z})$ 
because of the non-negativity of Kullback-Leibler divergence, and 
becomes tight only when the trial distribution $q(\mathbf{z})$ corresponds to the posterior $p({\mathbf{z}}|D,\theta )$.
This means $q(\mathbf{z})$ should be close to the posterior $p({\mathbf{z}}|D,\theta )$ to minimize the gap.

Because $\theta $ depends only on the first term of Eq. (\ref{eq:lowerbound}), 
the maximization of the marginal likelihood with respect to $\theta $ for some fixed $q(\mathbf{z})$ is equivalent to
the maximization of the first term of  Eq. (\ref{eq:lowerbound}), i.e.,
\begin{align}
{\theta ^*} = \arg \mathop {\max }\limits_\theta  F(q({\mathbf{z}}),\theta ) 
= \arg \mathop {\max }\limits_\theta  \sum\nolimits_t {\sum\nolimits_{\mathbf{z}} 
{q({\mathbf{z}})\log p({\mathbf{y}_t}|{{\mathbf{x}}_t},{\mathbf{z}}, \theta )} } , \label{eq:lowerbound maximization}
\end{align}

The above optimization of the lower bound by the stochastic gradient descent leads
the dropout training algorithm explained in Section~\ref{sec:notation}.
Note that we need to randomly sample  both the input-output pair $\{\mathbf{x}_t, \mathbf{y}_t \}$
and the mask ${\mathbf{z}}$.
The sample of the mask variable ${\mathbf{z}}$ should obey the trial distribution $q(\mathbf{z})$.
To be identical to the dropout training explained in Section~\ref{sec:dropout training},
we need to assume $p({\mathbf{z}}) = \prod\nolimits_{i = 1}^m p^{z_i}(1-p)^{z_i}$ and $q(\mathbf{z}) = p(\mathbf{z})$.
In this case, the parameter $p$ corresponds to the dropout rate.
In this paper, we will say $q(\mathbf{z})$ as the dropout rate 
because $q(\mathbf{z})$ determines the dropout rate.

\subsection{Bayesian interpretation of the prediction after dropout training}
We can also interpret the output for the test input ${\mathbf{x}}^*$ after learning of $\theta $ from a Bayesian standpoint.
In a Bayesian framework, the output for the test input ${\mathbf{x}}^*$ should be inferred as the
expected value of the predictive distribution given as 
\begin{align}
\int {\mathbf{y}}p({\mathbf{y}}|{\mathbf{x}}^*, D, \theta ) d{\mathbf{y}} = 
\sum\nolimits_{{\mathbf{z}}} \int {\mathbf{y}} p({\mathbf{y}}|{\mathbf{x}}^*, {\mathbf{z}}, \theta )d{\mathbf{y}}
p({\mathbf{z}}|D, \theta ) , \label{eq:predictive distribution}
\end{align}
That is, the output is predicted by the weighted average of 
submodel predictions $\int {\mathbf{y}} p({\mathbf{y}}|{\mathbf{x}}^*, {\mathbf{z}}, \theta )d{\mathbf{y}}$.
However, the above calculation becomes often intractable since
the exact calculation of both the posterior itself and the expectation with respect to the posterior
needs the summation over ${\mathbf{z}}$ which has $2^m$ variations.
Therefore, consider to replace the posterior $p({\mathbf{z}}|D, \theta )$ with %
some tractable trial distribution $q({\mathbf{z}})$.
\begin{align}
\int {\mathbf{y}}p({\mathbf{y}}|{\mathbf{x}}^*, D, \theta ) d{\mathbf{y}} \approx  
\sum\nolimits_{{\mathbf{z}}} \int {\mathbf{y}} p({\mathbf{y}}|{\mathbf{x}}^*, {\mathbf{z}}, \theta )d{\mathbf{y}}
q({\mathbf{z}} ) \approx  \int {\mathbf{y}} p({\mathbf{y}}|{\mathbf{x}}^*, E_{q(\mathbf{z})}[{\mathbf{z}}], \theta )d{\mathbf{y}}, %
\label{eq:approximated predictive distribution}
\end{align} 
%of the posterior $p({\mathbf{z}}|D, \theta )$ itself %and the expectaion with respect to the posteriror
When we use independent $p(\mathbf{z})$ for $q({\mathbf{z}} )$,
%$q(\mathbf{z}) = p(\mathbf{z})$,
%and apply a stochastic gradneint descent,
we will have the same output explained in Section~\ref{sec:dropout prediction}.
%The expecation over $q(\mathbf{x})$ needs further approximation.
%The simple one is
%\begin{align}
%\int {\mathbf{y}}p({\mathbf{y}}|{\mathbf{x}}^*, D, \theta ) d{\mathbf{y}} \approx  
%\int {\mathbf{y}} p({\mathbf{y}}|{\mathbf{x}}^*, E_{q(\mathbf{z})}[{\mathbf{z}}], \theta )d{\mathbf{y}}, 
%\end{align} 
The last approximation in Eq.(\ref{eq:approximated predictive distribution}) 
is the approximation to the expectation over $q(\mathbf{z})$.
This may be replaced by a Gaussian approximation explained in Section~\ref{sec:dropout prediction}.
%and the sophisticated one is to approximate the sum of random variable 
%by a single Gaussian variable applying Lyapunov's central limit theorem
%as explained in Section~\ref{sec:dropout prediction}.

%This lower bound (\ref{eq:lowerbound}) is closely related with the EM algorithm.
%Actually, if we replace the parameter $\mathbf{z}$ and  the product of likelihood 
%$ \sum\nolimits_{\mathbf{z}} {\left( {\prod\nolimits_{t = 1}^T {p({\mathbf{y}_t}|{{\mathbf{x}}_t},{\mathbf{z}},\theta )} } \right)}$
%with a time-dependent hidden variable $\mathbf{z}_t$ and one-time likelihood 
%$p({\mathbf{y}_t}|{{\mathbf{x}}_t},{\mathbf{z}},\theta )$ respectively, 
%and take a summation with respect to the samples $t$ on both sides, 
%then we can see the exact match with the EM algorithm.
%That is, in E-step, the posteiror $p({\mathbf{z}}_t|{\mathbf{x}}_t, {\mathbf{y}}_t, \theta )$ 
%is calculated and the trial distribution $q({\mathbf{z}}_t)$ is set to be the posterior
%$p({\mathbf{z}}_t|{\mathbf{x}}_t, {\mathbf{y}}_t, \theta )$ so as to minimize the gap, 
%then the lower bound is maximized with respect to $\theta $ in M-step.
%Note that only the first term and the second term 
%if the prior $p({\mathbf{z}} )$ is denepdent on $\theta $ as $p({\mathbf{z}}|\theta )$
%in lower bound relates to the parameter $\theta $ and the rest of the terms can be safely excluded in the optimization.

% Maximum likelihood estimation vs Bayesian inference

\section{Bayesian dropout algorithm with variable dropout rate} \label{sec:optimization}
As apparent from the discussion in the preceding section, 
$q(\mathbf{z})$ should be close to the posterior $p({\mathbf{z}}|D, \theta )$
because it makes the lower bound much more tight, which means the cost function $F(q(\mathbf{z}), \theta)$ 
with respect to $\theta $ is presumably closer to the desirable cost function, 
marginal log-likelihood ${\log p(D|\theta )}$. 
Moreover, it enables to approximate the predictive distribution much more accurately
when we use $q(\mathbf{z})$ that better approximates the posterior $p({\mathbf{z}}|D, \theta )$.

The best $q(\mathbf{z})$ is the posterior $p({\mathbf{z}}|D, \theta )$ which,
however, cannot be solved in an analytical form in most cases.
So consider to optimize $q(\mathbf{z})$ in a certain parametric distribution family, 
$q(\mathbf{z}|\lambda)$ that is parameterized by $\lambda $.
%that approximates the posterior 
Then the best parameter $\lambda ^*$ is obtained as  % $q(\mathbf{z}|\lambda ^*)$
the one that maximizes the lower bound,% with respect to $q(\mathbf{z})$,
\begin{align}
%{q^*}({\mathbf{z}})
\lambda ^* = \arg \mathop {\min }\limits_{\lambda} KL[q({\mathbf{z}}|\lambda)|p({\mathbf{z}}, D, \theta )] 
= \arg \mathop {\max }\limits_{\lambda } F(q({\mathbf{z}}| \lambda), \theta ).
\end{align}

%The above optimization, however, cannot be solved in an analytical form in most cases,
%and optimize the paramter $\lambda$ by a numerical optimization method 
The above optimization problem could be solved by a numerical optimization method 
such as a stochastic gradient descent.

Overall, the entire Bayesian dropout algorithm would be as follows;

%\begin{itembox}[l]{ \bf Bayesian dropout for learning }
\fbox{\parbox{\textwidth}{\textbf{Bayesian dropout for learning }:
\begin{enumerate}
\item Set  $t=0$ and set an initial estimate for parameters $\theta _0$ and $\lambda _0$.  \par 
\item Pick a pair of sample $({\mathbf{x}}_t, {\mathbf{y}}_t)$ at random. \par 
\item Randomly set a mask $\mathbf{z}_t$ by determining every element independently 
         according to the dropout rate $q(\mathbf{z} | \lambda _t)$. \par
\item Update the parameter $\theta $ as  \par
\begin{eqnarray}
 \theta _{t+1} = \theta _t + \eta _t \frac{\partial \log p(\mathbf{y}_t|\mathbf{x}_t, \mathbf{z}_t, \theta)}{\partial \theta }.
 \nonumber 
\end{eqnarray} 
\item Update the parameter of the dropout rate $q(\mathbf{z} | \lambda _t)$ as  \par
\begin{align}
  \lambda _{t+1} = &  \lambda _t + \epsilon _t \left\{ {\frac{{\partial \log q({\mathbf{z}}_t|\lambda )}}{{\partial \lambda }}}
  \log p({\mathbf{y}_t}|{{\mathbf{x}}_t},{\mathbf{z}}_t,\theta ) \right. \nonumber  \\ %\par
 % \frac{\partial }{\partial \lambda } \{
%\sum\nolimits_{t} \sum\nolimits_{\mathbf{z}} q({\mathbf{z}} | \lambda)\log p({\mathbf{y}_t}|{{\mathbf{x}}_t},{\mathbf{z}},\theta )
& \left. + \delta _t\frac{\partial }{\partial \lambda } \left( \sum\nolimits_{\mathbf{z}} q({\mathbf{z}}| \lambda) \log p({\mathbf{z}})
- \sum\nolimits_{\mathbf{z}} {q({\mathbf{z}}| \lambda)\log q({\mathbf{z}}| \lambda)} \right) \right\},
%%\log p(\mathbf{y}_t|\mathbf{x}_t, \mathbf{z}_t, \theta) + 
%%\delta (\log q(\mathbf{z}_t | \lambda ) -\log p(\mathbf{z}_t))  \},
%%%\delta \log\frac{q(\mathbf{z}_t | \lambda )}{p(\mathbf{z}_t)} \right),
 \label{dropout rate optimization}
\end{align} 
where $\eta _t$ and $\epsilon _t$ determine step sizes, and should be decreased properly to assure the convergence
while $\delta _t$ is an inverse of the effective number of samples, which will be explained later.
\par
%In case of the training of the three-layered neural network for regression, the above parameter update would be \par
%\begin{eqnarray}
% \theta _{t+1} = \theta _t - \eta _t \frac{\partial ||{\mathbf{y}}_t - {\mathbf{y}}({\mathbf{x}}_t;Z^{(1)}_t, Z^{(2)}_t, \theta )||^2 _2}{\partial \theta }
% \nonumber 
%\end{eqnarray} \par%
%where $||\cdot ||_2$ denotes Euclid norm.\par
\item Increment $t$ and go to step 2 unless certain termination
condition is not satisfied.
\end{enumerate}
}}

The first term in the right hand side of Eq. (\ref{dropout rate optimization}) comes from
the first term of the derivative of $F(q(\mathbf{z}|\lambda), \theta )/T$,
$\frac{\partial }{\partial \lambda }\frac{1}{T}\sum\nolimits_{t=1}^T \sum\nolimits_{\mathbf{z}} 
q({\mathbf{z}})\log p({\mathbf{y}_t}|{{\mathbf{x}}_t},{\mathbf{z}},\theta )
\approx {{\mathrm{E}}_{q({\mathbf{z}}|\lambda)}}\left[ 
{E_{r({\mathbf{x}},{\mathbf{y}})}}\left[ 
\frac{{\partial \log q({\mathbf{z}}|\lambda)}}{{\partial \lambda }}{\log p({\mathbf{y}}|{\mathbf{x}},{\mathbf{z}},\theta )} \right] \right]$
where $E_{q({\mathbf{z}}|\lambda )}\left[ \cdot \right]$ and $E_{r({\mathbf{x}},{\mathbf{y}})}\left[ \cdot \right]$
denote the expectation with respect to the trial distribution $q({\mathbf{z}}|\lambda )$ and the
true distribution of the training set $\{ {\mathbf{x}}_t,{\mathbf{y}}_t | t=1, \cdots , T \}$.
Because these two expectations cannot be evaluated analytically in most cases due to the complex dependence of 
$\log p({\mathbf{y}}|{\mathbf{x}},{\mathbf{z}},\theta )$ on both $\{ {\mathbf{x}},{\mathbf{y}} \}$ and $\mathbf{z}$,
the expectations are replaced by random samples according to the recipe of the stochastic gradient descent.
In contrast, the rest of two terms corresponding to the derivatives of the last two terms of $F(q(\mathbf{z}|\lambda), \theta )/T$
can be evaluated directly if we assume independent distributions for both 
$p({\mathbf{z}})=\prod \nolimits_{i=1}^m p(z_i)$ and $q({\mathbf{z}}) = \prod \nolimits_{i=1}^m q(z_i|\lambda )$.
The summation over $\mathbf{z}$ is usually intractable because
$\mathbf{z}$ has $2^m$ variations.
However, the independence assumption decomposes the intractable summation over $\mathbf{z}$ 
into $m$ tractable summations over a binary $z_i$ $(i=1,\cdots ,m)$.
This allows us the direct evaluation of these terms in practice without using Monte Carlo method.
%When the last two terms are evaluated directly, it is not necessary to evaluate them by a stochastic gradient descent.
That is the reason why the last two terms does not depend on the sample 
${\mathbf{z}}_t$ as well as $\{ {\mathbf{x}}_t, {\mathbf{y}}_t \}$.

A scalar $\delta _t$ denotes an inverse of the effective number of samples, i.e., $\delta _t \propto 1/T$
so that the last two terms correspond to the derivative of the last two terms of $F(q(\mathbf{z}|\lambda), \theta )/T$.
%However, this fixed amount data assumption is not compatible with the online learning scenario using stochastic gradient descent.
%An alternative choice would be $\delta _t \propto  1/t$ when we think the amount of the training data
%is increasing as the algorithm proceeds.
This may be scheduled as $\delta _t \propto  1/t$ when we think the amount of the training data
is increasing as the algorithm proceeds.
It is noted that many other variants of this algorithm can be considered.
For example, we can introduce momentum term, we can update the parameters for a mini-batch, 
or we can consider EM-like algorithm, i.e., the update of $\theta $ (steps 2-4) is repeatedly performed
until $\frac{\partial F(q(\mathbf{z}| \lambda ), \theta )}{\partial \theta } \approx 0$ holds, 
then proceeds to the update of $\lambda $, and $\lambda $ is updated 
until $\frac{\partial F(q(\mathbf{z}| \lambda ), \theta )}{\partial \lambda  } \approx 0$ holds.

Following the standard dropout training, one natural choice of the parameterization of 
 $q(\mathbf{z}|\lambda )$ would be $q(\mathbf{z}|\lambda ) 
 = {\prod\nolimits_{i = 1}^m {{\lambda^{z_i}}{{(1 - \lambda)}^{(1 - z_i)}}} }$
 where $\lambda $ represents a common dropout rate to all the input features.
 Hereafter, we will refer the dropout with the optimized uniform scalar $\lambda $ to uniformly optimized rate dropout (UOR dropout).
We could consider other parameterizations of $q(\mathbf{z}|\lambda )$.
For example, we can parameterize for each $q(z_i=1)$ differently by letting $q(z_i=1) = \lambda _i $.
 Hereafter, we will refer the dropout with the optimized element-wise $\lambda _i$ %
 to feature-wise optimized rate dropout (FOR dropout).
The parameter of the mask distribution can specify the distribution of subset of ${\mathbf{z}}$ independently, such as,
$q(z_j=1) = \lambda _i ({\rm{if \hspace{1pt}}} {\mathit{z_j}} \in I_i)$ where $I_i$ denotes $i$-th subset of ${\mathbf{z}}$.
The subset could be the units in a same layer.
This layer-wise setting of the dropout rate is sometimes used in practice, e.g., ~\citep{Graham14spatiallyCNN}.

\section{Experiment}  \label{sec:experiment}
We test the validity of the optimization of the dropout rate
with a binary classification problem.

\subsection{Data}
Let $y \in \{0, 1 \}$ be a target binary label and ${\mathbf{x}}$ be a 
1000-dimensional input vector consisting of 100-dimensional informative features 
and the rest of 900-dimensional non-informative features.
Label $y $ obeys ${\rm{Ber}}(0.5)$.
Each informative feature $x_i$ ($i=1,\cdots ,100$) is generated independently according to $p(x_i|y=0)=N(x_i|-0.1, 1)$
and $p(x_i|y=1)=N(x_i|0.1, 1)$ so that 
each informative feature has a weak correlation with the label $y$ (cross-correlation is 0.1)
while  each non-informative feature  $x_i$ ($i=101,\cdots ,1000$) is generated independently according to $p(x_i)=N(x_i|0, 1)$
irrelevant to the label $y$.
Here, $N(x|\mu, s^2)$ denotes a Gaussian distribution with mean $\mu$ and variance $s ^2$.
2000 samples are generated for the training, and another 1000 samples are generated for the validation. 
The performance is evaluated with another 20,000 test samples.

\subsection{Model and Algorithm }
Linear logistic regression is used for this task.
\begin{align}
p({\mathbf{y}} = 1|{\mathbf{x}},{\mathbf{z}},\theta ) = \frac{1}{{1 + \exp ( - {\theta ^T}Z{\mathbf{x}})}} \equiv \sigma ({\theta ^T}Z\mathbf{x}),
\end{align}
where $Z$ is a diagonal matrix whose diagonal elements are binary masks ${\mathbf{z}}$.
In this case, the Gaussian approximation to ${\theta ^T}Z{\mathbf{x}}$ works effectively %
as proposed in ~\citep{Wang13fastdropout}.
Approximating $u = {\theta ^T}Z{\mathbf{x}}$ by a Gaussian random variable %$N(x_i|\mu , s^2)$ where 
whose mean is $\mu = \sum \nolimits_{i=1}^m q(z_i=1)\theta _i x_i$ and whose variance is
$s^2 = q(z_i=0)q(z_i=1)(\theta _i x_i)^2 $, and applying a well-known formula which %
approximates Gaussian integral of sigmoid function $\sigma (\mathbf{x})$, the predictive distribution is obtained as
\begin{align}
p({\mathbf{y}} = 1|{\mathbf{x}}, \theta ) \approx \int {\sigma(u)}N(u|\mu,s^2)du 
\approx \sigma \left( \frac{\mu }{\sqrt{1+\pi s^2/8}}  \right).
\end{align}

We compared four algorithms, 1) maximum likelihood estimation (MLE), 2) standard dropout algorithm (fixed dropout) 
where the dropout rate is fixed to be 0.5,
3) Bayesian dropout algorithm that optimizes a uniform dropout rate (UOR dropout), and
4) Bayesian dropout algorithm that optimizes feature-wise dropout rates (DOR dropout).
%The trial distribution $q(\mathbf{z}|\lambda )$ is assumed to be independent, $q(\mathbf{z}|\lambda ) = \prod _i q(z_i| \lambda )$,
%and each component distribution $q(z_i| \lambda )$ is parametrized as $q(z_i=1| \lambda ) = \sigma(\lambda ) $ 
%for UOR dropout, and $q(z_i=1| \lambda ) = \sigma(\lambda _i) $ for all adaptive dropout.
For the optimization of $\theta $ and $\lambda $, we need to determine the step sizes $\eta _t$
and $\epsilon _t$ of the algorithm described in Section~\ref{sec:optimization} properly. %
These values are scheduled as $\eta _t = \frac{a}{1+t/b} $ and $\epsilon _t = \frac{c}{1+t/d}$.
The step size $\epsilon _t$ used for updating $\lambda  $ should be decreased slower
 than $\eta $ used for updating $\theta$.
Considering this, we chose the best parameter set $a, b, c$, and $d$ 
that shows the best accuracy for the validation data among
 $a \in \{3 \cdot 10^{-4}, 10^{-3}, 3 \cdot 10^{-2}, 10^{-2}\}$, 
 $b \in \{10^{2}, 10^{3}, 10^{4} \}$, 
 $c \in \{3 \cdot 10^{-4}, 10^{-3}, 3 \cdot 10^{-2}, 10^{-2}\}$, 
 $d \in \{10^{3}, 10^{4}, 10^{5} \}$, respectively.
For Bayesian dropout algorithm, the inverse of the effective number of samples $\delta $ is %
simply set to be the inverse of the number of samples, i.e., $10^{-3}$,
and the initial value of the dropout rate is set to be all 0.5 for both UOR dropout and DOR dropout.

\subsection{Result}
Test accuracy after learning is shown in Fig.1(a),
and the trained dropout rate is shown in Fig.1(b).
In Fig.1(b), the height of the bar denotes the dropout rate of FOR dropout
while '+' denotes the dropout rate of UOR dropout.

%%%%%%%%%%%%%%%%%%%%%%%%%%%%%%%%%%%%%%%%%%%%%%%%%%%%%
\begin{figure}%[h] % t:page top, h: here, b: page bottom
\centering
\includegraphics[width =14cm]{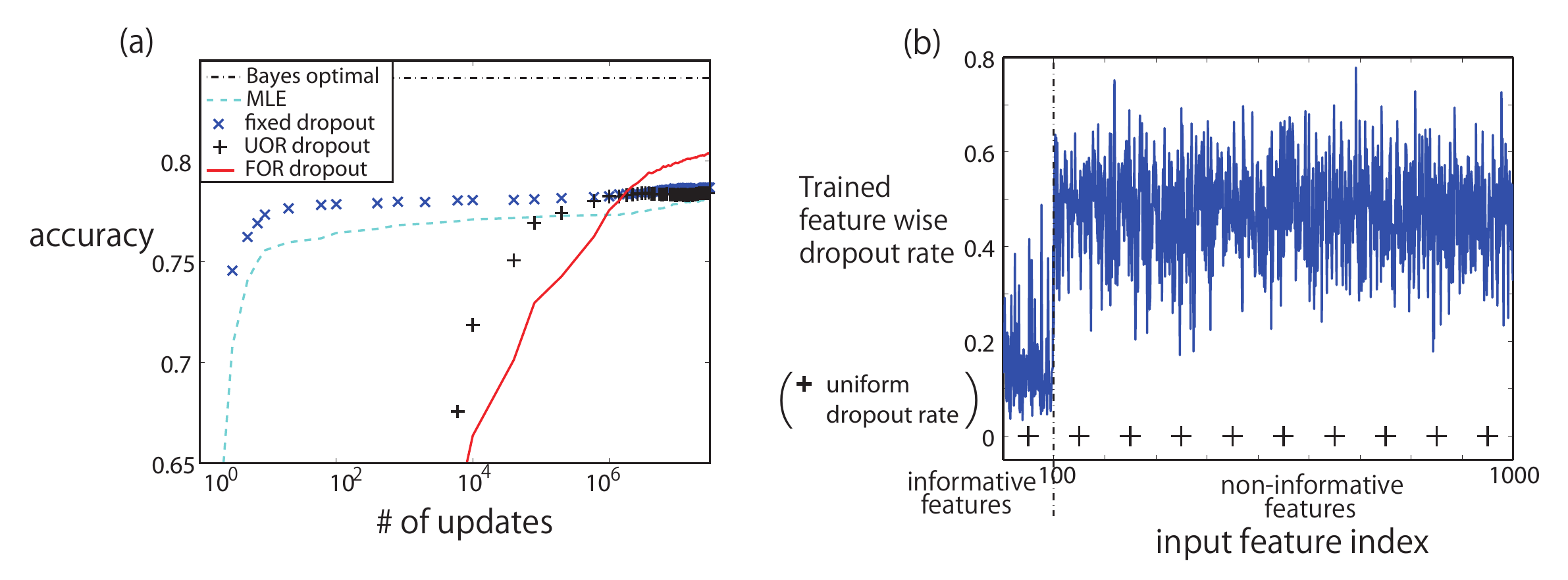} %{RBM.pdf} 
\caption{Experimental results (a) Test accuracy \quad (b) Dropout rate after learning}
\label{fig1}
%\begin{minipage}{0.96\linewidth}
%\quad \quad \quad \quad \quad \quad \quad \quad \quad \quad \quad \quad '+' denotes the dropout rate of single adaptive dropout.
%\end{minipage}
\end{figure}
%%%%%%%%%%%%%%%%%%%%%%%%%%%%%%%%%%%%%%%%%%%%%%%%%%%%%%
%%%%%%%%%%%%%%%%%%%%%%%%%%%%%%%%%%%%%%%%%%%%%%%%%%%%%%
%\begin{figure} % t:page top, h: here, b: page bottom
%\centering
%\includegraphics[width=0.4\linewidth]{test_accuracy_Fig1.pdf} %{RBM.pdf} 
%\caption{Comparison of test accuracy}
%\label{fig1}
%\begin{minipage}{0.96\linewidth}
%\end{minipage}
%\end{figure}
%%%%%%%%%%%%%%%%%%%%%%%%%%%%%%%%%%%%%%%%%%%%%%%%%%%%%%
%%%%%%%%%%%%%%%%%%%%%%%%%%%%%%%%%%%%%%%%%%%%%%%%%%%%%%
%\begin{figure} % t:page top, h: here, b: page bottom
%\centering
%\includegraphics[width=0.4\linewidth]{optimized_q_Fig2.pdf} %{RBM.pdf} 
%\caption{Optimized dropout rate of all adaptive dropout}
%\label{fig2}
%\begin{minipage}{0.96\linewidth}
%'+' denotes the dropout rate of single daptive dropout.
%\end{minipage}
%\end{figure}
%%%%%%%%%%%%%%%%%%%%%%%%%%%%%%%%%%%%%%%%%%%%%%%%%%%%%%
Because only 10$\%$ of features are relevant to the input,
it is difficult to determine the best uniform dropout rate.
Eventually, the optimized uniform dropout rate becomes zero.
Then, the test accuracies of both MLE and UOR dropout takes almost same value.
Fixed dropout rate shows a slightly better accuracy although the difference 
cannot be visible from Fig. 1(a).
%fixed dropout rate and 
%UOR dropout rate takes a value close to the one obtained by MLE.
On the other hand, if we optimize the feature-wise dropout rate,
the test accuracy increases, getting closer to the Bayes optimal (theoretical limit).
As can be seen from Fig1.(b), 
only the dropout rate of the informative features (the first 100 elements)
are selectively low. 
Note that FOR dropout shows a significant regularization effect
although it has no hyperparameter to be tuned.

\section{Discussion} \label{sec:discussion}
\subsection{Why Bayesian?} 
The input feature selection problem and the hidden internal units selection problem is, %
in general, difficult because it requires to compare a vast number of models as large as
$2^n$ and $2^m$ in case of the three-layered neural network explained in Section~\ref{sec:notation}.

Moreover, the maximum likelihood criterion is not useful in choosing the best mask $\mathbf{z}$
because it always prefers the largest network, that is, it always chooses $\mathbf{z}=\mathbf{1}$
where $\mathbf{1}$ is a vector containing all 1 in its elements. 
\begin{align}
\mathop {\max }\limits_{{\mathbf{z}}, \theta } \sum\nolimits_t {\log p({\mathbf{y}_t}|{{\mathbf{x}}_t},{\mathbf{z}}, \theta )} 
= \mathop {\max }\limits_\theta  \sum\nolimits_t {\log p({\mathbf{y}_t}|{{\mathbf{x}}_t},{\mathbf{z}} = {\mathbf{1}}, \theta )}.
\end{align}
It is a well known fact that the number of the parameters has a proportional relation %
with the discrepancy between the generalization error %
and the training error of the non-singular statistical models trained by 
$\it{Maximum}$ $\it{a}$ $\it{Posteriori}$ (MAP) estimation which uses a fixed prior 
and the maximum likelihood estimation (see Section 6.4 of \citet{Watanabe09AlgebraicGeoemtry} for example).

In contrast, Bayesian inference does not rely on the prediction of a single model,
but on the weighted sum of submodel predictions where %
the weights are determined according to the posterior of the submodel.%
The marginal likelihood integrated over all possible submodels can
take into account the complexity of the model. 
This fact helps to prevent the overfitting of the type explained in Section 3.4 of ~\citep{Bishop06PRML}.
As for the asymptotic behavior of the Bayes generalization error,
there are studies from algebraic geometry~\citep{Watanabe09AlgebraicGeoemtry},
which reveal that the asymptotic behavior of a singular statistical model is different from
that of the non-singular statistical model; 
The generalization error does not increase necessary proportionally to the number of the parameters.
This partly explains the reason why
%Maybe we could find a reason from this theoretical study %
the recent big neural network, which is categorized into a singular statistical model, can resist overfitting
as opposed to the regular statistical model optimized by the maximum likelihood estimation or MAP estimation.

\subsection{What's different from the other studies?} 
%As far as the authors know, 
There are several studies that view dropout training %
as a kind of approximate Bayesian learning except for the original study~\citep{Hinton12dropout}.
However, their interpretation differ from our interpretation or lacks solid theoretical foundation.

% as explained in~Section\ref{sec:existing methods}.%

In~\citep{Wang13fastdropout}, it is stated that the cost function for $\theta $ 
can be seen as the lower bound of the marginal log-likelihood. %
However, their marginal log-likelihood is defined as %
$\log [E_{p(\mathbf{z})} p(\mathbf{y}|\theta ^T Z\mathbf{x})]$
which has a lower bound $E_{p(\mathbf{z})}[\log p(\mathbf{y}|\theta ^T Z\mathbf{x})]$
with an independent Bernoulli distribution $p(\mathbf{z})$ as the dropout rate. %
%  which determines the dropout rate.
Similar cost function %$E_{q(\mathbf{z}|\mathbf{x}, \lambda )}[\log p(\mathbf{y}|\theta ^T Z\mathbf{x})]$%
 is proposed by~\citep{Ba13regularizeddropout}.
They propose to optimize the dropout rate $q(\mathbf{z}|\mathbf{x}, \lambda )$ %
so as to maximize $E_{q(\mathbf{z}|\mathbf{x}, \lambda )}[\log p(\mathbf{y}|\theta ^T Z\mathbf{x})]$.
Basically, their view on $\mathbf{z}$ is a hidden variable rather than a parameter
because $\mathbf{z}$ is not inferred by the entire training dataset,
but inferred sample by sample as $p(\mathbf {z}_t|\mathbf {x}_t, \mathbf {y}_t)$.
%which can be derived from the complete likelihood $p(\mathbf {y}_t, \mathbf {z}_t|\mathbf {x}_t)$.
In contrast, we define the marginal likelihood as 
$ {\log p(D|\theta )} \equiv  \log \sum\nolimits_{\mathbf{z}} {\left( {\prod\nolimits_{t = 1}^T 
{p({\mathbf{y}_t}|{{\mathbf{x}}_t},{\mathbf{z}},\theta )} } \right)}p({\mathbf{z}})$.
With this definition, we may interpret the computation of the
expectation of the submodel prediction with respect to the submodel posterior  
$p(\mathbf{z}|D, \theta )$  (i.e. Bayesian model averaging) 
as a Bayesian way to solve the model selection problem.
%This formulation allows us to interpret the computation of the expectation with respect to approximation of %
%the posterior $p(\mathbf{z}|D, \theta )$ as the approximate Bayesian model averaging 
%that solves the input feature selection, hidden internal units selection or their combination in a Bayesian way.%
%Approximation of the true posterior $p(\mathbf{z}|D, \theta )$ by a independent distribution 
%$q({\mathbf{z}}) = \prod _i q(z_i)$ eases the original intractable problem
%although it is still time-consuming.
%Here, we show the effectiveness of the optimization of the dropout rate %
%by an artificail classification problem.%
%Good parameterization will make the optimization much easer.

We shall note that, in the context of the neural network, there is nothing new about  
the philosophy of assigning the prior to the parameter $\theta$
and seeking the posterior achieving the best Bayesian prediction Eq. (\ref{eq:predictive distribution}). 
See, for example, ~\citep{Neal96BayesNN, Xiong11BayesNN}, which assigned 
Gaussian distribution to the weight matrices $W$s. 
The search of the optimal prior over the vast space of probability measures is indeed a daunting task, however.
The aforementioned algorithms do suffer from runtime.  
From this computational point of view, the standard dropout algorithm emerges as an efficient Bayesian compromise to this search. 
In particular, it restricts the search of the distribution of each $(i,j)$ element of $W$, $W_{ij}$ 
to the ones for which one can write as 
$$W_{ij}= z_{j} \tilde W_{ij}    ~~~~~ z_j \sim {\rm{Ber}}(0.5)~ iid,   ~~~ \tilde W_{ij} ~~~\textrm{deterministic}$$
for all $j$-th unit adjacent to the connection $(i,j)$
\footnote{To precisely correspond to the original dropout algorithm, %
we also need to set the prior $p(\mathbf{z})$ appropriately %
so that the reguralization term
$\sum\nolimits_{\mathbf{z}} q({\mathbf{z}}) \log p({\mathbf{z}})  - \sum\nolimits_{\mathbf{z}} {q({\mathbf{z}})\log q({\mathbf{z}})}$
becomes identical with the reguralization term used in the original dropout aglorithm.}. 
%we also need a condition that the prior for $W_{ij}$, 
%$p(W_{ij} )$ is non-informative, i.e., uniform distribution within the range of $W_{ij}$.
In other words, the algorithm only looks into a specific family of discrete valued distributions parametrized by the values of $\tilde W_{ij}$.
This way, the standard dropout algorithm bridges between the stochastic feature selection and the Beyesian averaging.\\

In this light, our algorithm can be seen as one step improvement of the standard dropout algorithm, 
which restricts the search of the posterior distribution of $W_{ij}$ to the ones of the form 
$$W_{ij}= z_j \tilde W_{ij}    ~~~~~ z_j \sim {\rm{Ber}}(p_j)~ iid,   ~~~ \tilde W_{ij} ~~~\textrm{deterministic}$$
for all $j$-th unit adjacent to the connection $(i,j)$. 
This is a set of discrete valued distributions parametrized by $\tilde W_{ij}$ and $p_j$. 
Because the search space becomes larger, our method will consume more runtime than the standard dropout algorithm.  
By allowing more freedom to the distribution, however, one can expect the trained machine to be much more data specific. 
As apparent from the above discussion, we can consider the dropconnect~\citep{Wan13DropConnect} 
as another parameterization of the distribution of $W_{ij}$,
i.e., $W_{ij}= z_{ij} \tilde W_{ij}$ where $z_{ij}$ obeys a Bernoulli distribution.

%It also should be noted that the Bayesian model averaging is performed 
%in other studies such as Bayesian neural network~\citep{Neal96BayesNN, Xiong11BayesNN}.
%However, the dropout takes somewhat unique approach to the Bayesian model averaging
%% taken in dropout is different from the one 
%%in Bayesian neural network~\citep{Neal96BayesNN, Xiong11BayesNN}
%in the sense that dropout treats the weights of the neural network 
%not as the random variables, but as the deterministic parameters, 
%and the binary mask variable $\mathbf{z}$ is treated as the random variable to be inferred.
%This change connects the Bayesian model averaging
%with the discrete feature selection problem. 
%Also, from a computation aspect, the dropout takes advantage of %
%the simplification of the computation by the drastic parameterization.
%We can think that the standard dropout algorithm utilizes
%the simplest parametric distribution, i.e., independent Bernoulli 
%distribution with a fixed probability 0.5.
%In this study, we propose to extend this parametrization a little bit %
%more broader parametric family, e.g., independent Bernoulli distribution 
%with a uniform probability or independent Bernoulli distribution having %
%feature-wise different probabilities at the expense of the
%increase of the computation time.
%In the experiment, we show the performance gain by this optimziation %
%is worth the expense of the computation time for some task.
%In general, it is desirable to consider the task-specific parameterization.
If need be, one can also group $p_j$s to match the specific task required for the model.
For example, let us consider the family of time-series prediction model such as vector autoregression (VAR) model:
$$X_t= \sum_{k=1}^M A_kX_{t-k} ~~~~~ X \in \mathbb{R}^d.$$  
For this family, the number of parameters will grow quickly with the state space dimension $d$ and the size of lag time $M$, making the 
search for the best posterior distribution difficult. Again, we may put $A_k \sim Z^{(k)} \tilde A_k $ where $Z^{(k)}$ are diagonal matrices with entries 
$\{ {\rm{Ber}}(p^{(k)}_i) ~~ i =1, \cdots ,d\}$. We may reduce the number of the hyperparameters 
by assuming some structure to the set of $p^{(k)}_i$s; 
for example, we can put $p^{(k)}_i = \tilde{p}_k$ or $p^{(k)}_i = \tilde{p}_k \lambda_i$ and introduce an intended space-time correlations. 
%In this case, it may be useful to set the different probability for each time-delay
%so that the difference of the temporal correlation can be represented, or %
%set the different probability to the same element or the others.
This type of parametrization % inspired by dropout 
may allow us to resolve a very sophisticated feature selection problems, which are considered intractable %
by the conventional methods.

%We can extend this 
%just 
%at the expense of the computation time.
%
%As done in the otiginal paper~\citep{Hinton12dropout}
%of the distribution of the binary variable.
%In approximate Bayesian inference, it is common to approximate the posterior 
% by certain indenpendent distribution to ease the computation.
% In dropout, not only the independence assumption but also the parametrization
% make use of the parametrization.
%This binary posterior is greatly simplified by a parametrization.
%For example, the optimization of a single dropout rate is just an optimization of a scalar value.
%%, but dropout modifies the learning of the original model parameters
%%indirectly through randomly masking the input and hidden units 
%%where the masks can be considered as a binary random variables
%On the other hand, Bayesian neural network treats all the network weights 
%as continuous random variables to be inferred in a Bayesian way.
%Therefore, the training of  Bayesian neural network is, in general, time-consuming,
%which prevents the application of Bayesian neural network 
% to a large scale of the data and network.

%\bibliography{iclr2015}
\bibliography{item2}
\bibliographystyle{iclr2015}

\end{document}